%% file: main.tex
\documentclass[runningheads]{llncs}

\usepackage{hyperref}
\usepackage{graphicx}
\usepackage{amsmath,mathtools}
\usepackage{amssymb}
\usepackage{xspace}
\usepackage{verbatim}
\usepackage[ruled,vlined]{algorithm2e}
\usepackage{algorithmic}

\title{Explaining Trained Neural Networks with Semantic Web Technologies: First Steps} 
\titlerunning{Explaining Trained Neural Networks with Semantic Web Technologies}

\author{Md Kamruzzaman Sarker \and Ning Xie \and Derek Doran \and Michael Raymer \and Pascal Hitzler}
\institute{Data Science and Security Cluster, Wright State University, Dayton, OH, USA }

\authorrunning{Sarker, Xie, Doran, Raymer, Hitzler}

\begin{document}

\maketitle

\input{abstract}
\input{intro}
\input{preliminaries}
\input{newresults}

\input{conclusion}

\bigskip

\noindent\emph{Acknowledgements.} This work was supported by the Ohio Federal Research Network project \emph{Human-Centered Big Data}.

\urlstyle{same}
\bibliographystyle{splncs03}
\bibliography{all}



\end{document}

%% file: abstract.tex
\begin{abstract}
The ever increasing prevalence of publicly available structured data on the World Wide Web enables new applications in a variety of domains. In this paper, we provide a conceptual approach that leverages such data in order to explain the input-output behavior of trained artificial neural networks. We apply existing Semantic Web technologies in order to provide an experimental proof of concept. 
\end{abstract}


%% file: intro.tex
\section{Introduction}\label{sec:intro}


Trained neural networks are usually imagined as black boxes, in that they do not give any direct indications why an output (e.g., a prediction) was made by the network. The reason for this lies in the distributed nature of the information encoded in the weighted connections of the network. Of course, for applications, e.g., safety-critical ones, this is an unsatisfactory situation. Methods are therefore sought to explain how the output of trained neural networks
are reached. 

This topic of explaining trained neural networks is not a new one, in fact there is already quite a bit of tradition and literature on the topic of rule extraction from such networks (see, e.g., \cite{BaderH05,AZ99,LehmannBH10}), which pursued very similar goals. Rule extraction, however, utilized propositional rules as target logic for generating explanations, and as such remained very limited in terms of explanations which are human-understandable. Novel deep learning architectures attempt
to retrieve explanations as well, but often the use-case is only for computer vision tasks like object
 or scene recognition. Moreover, explanations in this context actually encode greater details about
 the images provided as input, rather than explaining why or how the neural network was able to 
 recognize a particular object or scene. 
 
Semantic Web \cite{tbl-semantic-web,FOST} is concerned with data sharing, discovery, integration, and reuse. As field, it does not only target data on the World Wide Web, but its methods are also applicable to knowledge management and other tasks off the Web. Central to the field is the use of knowledge graphs (usually expressed using the W3C standard Resource Description Framework RDF \cite{turtle}) and type logics attached to these graphs, which are called \emph{ontologies} and are usually expressed using the W3C standard Web Ontology Language OWL \cite{owl2-primer}. 

This paper introduces a new paradigm for explaining neural network behavior. It goes beyond the limited propositional paradigm, and directly targets the problem of explaining neural network activity 
rather than the qualities of the input. The paradigm leverages advances in knowledge representation on the World Wide Web, more precisely from the field of Semantic Web technologies. 
It in particular utilizes the fact that methods, tool, and structured data in the mentioned formats are now widely available, and that the amount of such structured data on the Web is in fact constantly growing \cite{lod2009,LehmannIJJKMHMK15}. Prominent examples of large-scale datasets include Wikidata \cite{VrandecicK14} and data coming from the schema.org \cite{GuhaBM16} effort which is driven by major Web search engine providers. We will utilize this available data as background knowledge, on the hypothesis that background knowledge will make it possible to obtain more concise explanations. This addresses the issue in propositional rule extraction that extracted rulesets are often large and complex, and due to their sizes difficult to understand for humans. While the paper only attempts to explain input-output behavior,
the authors are actively exploring ways to also explain internal node activations. 

\subsection*{An illustrative example}
Let us consider the following very simple example which is taken from \cite{nesy17-prop}. Assume that the input-output mapping $P$ of the neural network without background knowledge could be extracted as
$$
p_1 \wedge q \to r \qquad
p_2 \wedge q \to r.
$$
Now assume furthermore that we also have background knowledge $K$ in form of the rules
$$
p_1 \to p\qquad
p_2 \to p.
$$
The background knowledge then makes it possible to obtain the simplified input-output mapping $P_K$, as
$$p \wedge q \to r.$$ 

The simplification through the background knowledge is caused by $p$ acting as a ``generalization'' of both $p_1$ and $p_2$. For the rest of the paper it may be beneficial to think of $p$, $p_1$ and $p_2$ as classes or concepts, which are hierarchically related, e.g., $p_1$ being ``oak,'' $p_2$ being ``maple,'' and $p$ being ``tree.'' 

Yet this example is confined to propositional logic.\footnote{How to go beyond the propositional paradigm in neural-symbolic integration is one of the major challenges in the field \cite{AAAISS15}.} In the following, we show how we can bring structured (non-propositional) Semantic Web background knowledge to bear on the problem of explanation generation for trained neural networks, and how we can utilize Semantic Web technologies in order to generate non-propositional explanations. This work is at a very early stage, i.e., we will only present the conceptual architecture of the approach and minimal experimental results which are encouraging for continuing the effort.

The rest of the paper is structured as follows. In Section \ref{sec:prelim} we introduce notation as needed, in particular regarding description logics which underly the OWL standard, and briefly introduce the DL-Learner tool which features prominently in our approach. In Section~\ref{sec:newresults} we present the conceptual and ex\-peri\-men\-tal setup for our approach, and report on some first experiments. In Section \ref{sec:conclusion} we conclude and discuss avenues for future work.

%% file: preliminaries.tex
\section{Preliminaries}\label{sec:prelim}


We describe a minimum of preliminary notions and information needed in order to keep this paper relatively self-contained. \emph{Description logics} \cite{BCMNP07,FOST} are a major paradigm in knowledge representation as a subfield of artificial intelligence. At the same time, they play a very prominent role in the Semantic Web field since they are the foundation for one of the central Semantic Web standards, namely the W3C Web Ontology Language OWL \cite{owl2-primer,FOST}. 

Technically speaking, a description logic is a decidable fragment of first-order predicate logic (sometimes with equality or other extensions) using only unary and binary predicates. The unary predicates are called \emph{atomic classes},\footnote{or \emph{atomic concepts}} while the binary ones are refered to as \emph{roles},\footnote{or \emph{properties}} and constants are refered to as \emph{individuals}. In the following, we formally define the fundamental description logic known as $\mathcal{ALC}$, which will suffice for this paper. OWL is a proper superset of $\mathcal{ALC}$. 

Desciption logics allow for a simplified syntax (compared to first-order predicate logic), and we will introduce $\mathcal{ALC}$ in this simplified syntax. A translation into first-order predicate logic will be provided further below. 

Let $\mathcal{C}$ be a finite set of atomic classes, $\mathcal{R}$ be a finite set of roles, and $\mathcal{N}$ be a finite set of individuals. Then \emph{class expressions} (or simply, \emph{classes}) are defined recursively using the following grammar, where $A$ denotes atomic classes from $\mathcal{A}$ and $R$ denotes roles from $\mathcal{R}$. The symbols $\sqcap$ and $\sqcup$ denote conjunction and disjunction, respectively.
$$C,D ::= A\mid\lnot C\mid C\sqcap D\mid C\sqcup D\mid\forall
R.C\mid\exists R.C$$

A \emph{TBox} is a set of statements, called (\emph{general class inclusion}) \emph{axioms}, of the form $C\sqsubseteq D$, where $C$ and $D$ are class expressions -- the symbol $\sqsubseteq$ can be understood as a type of subset inclusion, or alternatively, as a logical implication. An \emph{ABox} is a set of statements of the forms $A(a)$ or $R(a,b)$, where $A$ is an atomic class, $R$ is a role, and $a,b$ are individuals.  A description logic \emph{knowledge base} consists of a TBox and an ABox. The notion of \emph{ontology} is used in different ways in the literature; sometimes it is used as equivalent to TBox, sometimes as equivalent to knowledge base. We will adopt the latter usage. 

We characterize the semantics of $\mathcal{ALC}$ knowledge bases by giving a translation into first-order predicate logic. If $\alpha$ is a TBox axiom of the form $C\sqsubseteq D$, then $\pi(\alpha)$ is defined inductively as in Figure~\ref{fig:ALCtoFOL}, where $A$ is a class name. ABox axioms remain unchanged. 

\begin{figure}[t]
\begin{align*}
\pi(C\sqsubseteq D)&= (\forall x_0)(\pi_{x_0}(C)\to\pi_{x_0}(D))\\
\pi_{x_i}(A)&=A(x_i)\\
\pi_{x_i}(\lnot C)&= \lnot\pi_{x_i}(C)\\
\pi_{x_i}(C\sqcap D)&= \pi_{x_i}(C)\wedge\pi_{x_i}(D)\\
\pi_{x_i}(C\sqcup D)&= \pi_{x_i}(C)\vee\pi_{x_i}(D)\\
\pi_{x_i}(\forall R.C)&=(\forall x_{i+1})(R(x_i,x_{i+1})\to\pi_{x_{i+1}}(C))\\
\pi_{x_i}(\exists R.C)&=(\exists x_{i+1})(R(x_i,x_{i+1})\wedge\pi_{x_{i+1}}(C))
\end{align*}
\caption{Translating TBox axioms into first-order predicate logic. We use auxiliary functions $\pi_{x_i}$, where the $x_i$ are variables. The axiom $A\sqsubseteq\exists R.\exists S.B$, for example, would be translated to $(\forall x_0)((A(x_0))\rightarrow(\exists x_1)(R(x_0,x_1)\wedge (\exists x_2)(S(x_1,x_2)\wedge B(x_2))))$. }\label{fig:ALCtoFOL}
\end{figure}

DL-Learner \cite{BuhmannLW16,LehmannH10} is a machine learning system inspired by inductive logic programming \cite{MuggletonR94}. Given a knowledge base and two sets of individuals from the knowledge base -- called positive respectively negative examples -- DL-Learner attempts to construct class expressions such that all the positive examples are contained in each of the class expressions, while none of the negative examples is. DL-Learner gives preference to shorter solutions, and in the standard setting returns approximate solutions if no fully correct solution is found. The inner workings of DL-Learner will not matter for this paper, and we refer to \cite{BuhmannLW16,LehmannH10} for details. However, we exemplify its functionality by looking at Michalski's trains as an example, which is a symbolic machine learning task from \cite{LarsonM77}, and which was presented also in \cite{LehmannH10}. 

\begin{figure}[t]
\begin{center}
\includegraphics[width=.8\textwidth]{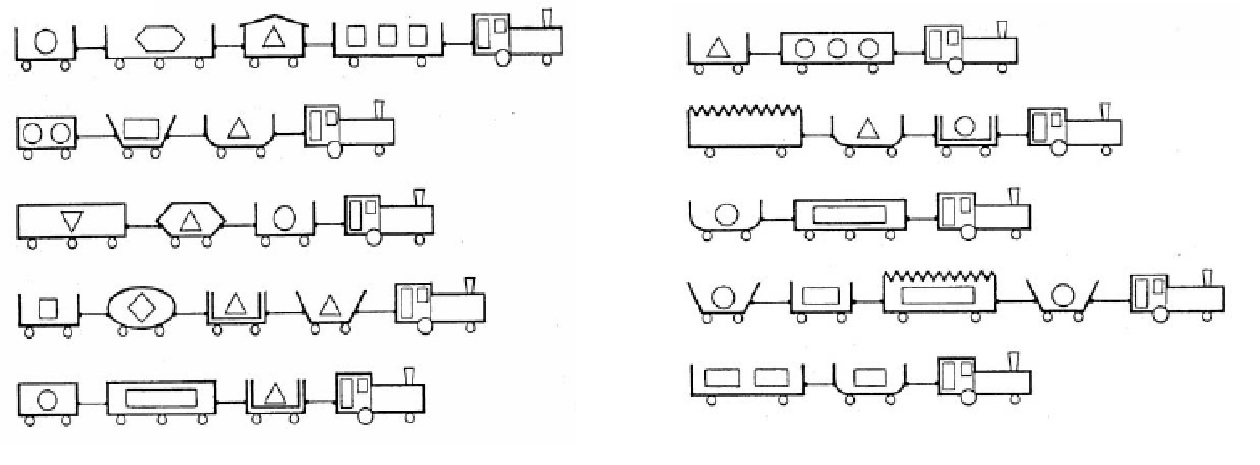}
\end{center}
\caption{Michalski's trains, picture from \cite{LarsonM77}. Positive examples on the left, negative ones on the right.}\label{fig:trains}
\end{figure}

For purposes of illustrating DL-Learner, Figure \ref{fig:trains} shows two sets of trains, the positive examples are on the left, the negative ones are on the right. Following \cite{LehmannH10}, we use a simple encoding of the trains as a knowledge base: Each train is an individual, and has cars attached to it using the hasCar property, and each car then falls into different categories, e.g., the top leftmost car would fall into the classes Open, Rectangular and Short, and would also have information attached to it regarding symbol carried (in this case, square), and how many of them (in this case, one). 
Given these examples and knowledge base, DL-Learner comes up with the class
$$\exists\text{hasCar}.(\text{Closed}\sqcap\text{Short})$$
which indeed is a simple class expression such that all positive examples fall under it, while no negative example does.

%% file: newresults.tex
\section{Approach and Experiments}\label{sec:newresults}

In this paper, we follow the lead of the propositional rule extraction work mentioned in the introduction, with the intent of improving on it in several ways.
\begin{enumerate}
\item We generalize the approach by going significantly beyond the propositional rule paradigm, by utilizing description logics. 
\item We include significantly sized and publicly available background knowledge in our approach in order to arrive at explanations which are more concise. 
\end{enumerate}

\begin{figure}[t]
\begin{center}
\includegraphics[width=\textwidth]{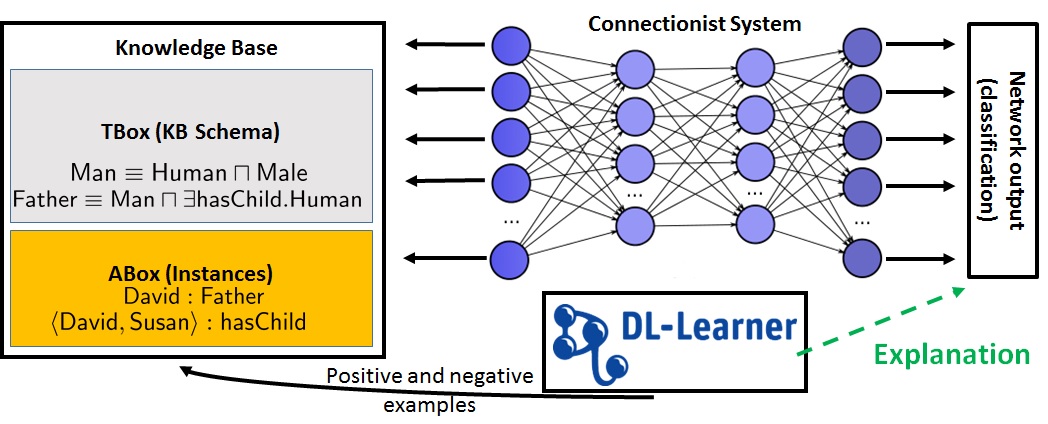}
\end{center}
\caption{Conceptual architecture -- see text for explanations.}\label{fig:overview}
\end{figure}

More concretely, we use DL-Learner as the key tool to arrive at the explanations. Figure \ref{fig:overview} depicts our conceptual architecture: The trained artificial neural network (connectionist system) acts as a classifier. Its inputs are mapped to a background knowledge base and according to the networks' classification, positive and negative examples are distinguished. DL-Learner is then run on the example sets and provides explanations for the classifications based on the background knowledge. 

In the following, we report on preliminary experiments we have conducted using our approach. Their sole purpose is to provide first and very preliminary insights into the feasibility of the proposed method. All experimental data is available from {\small \url{http://daselab.org/projects/human-centered-big-data}}.

\begin{figure}[t]
\begin{center}
\begin{minipage}{.45\textwidth}
\begin{center}
\includegraphics[height=.5\textwidth]{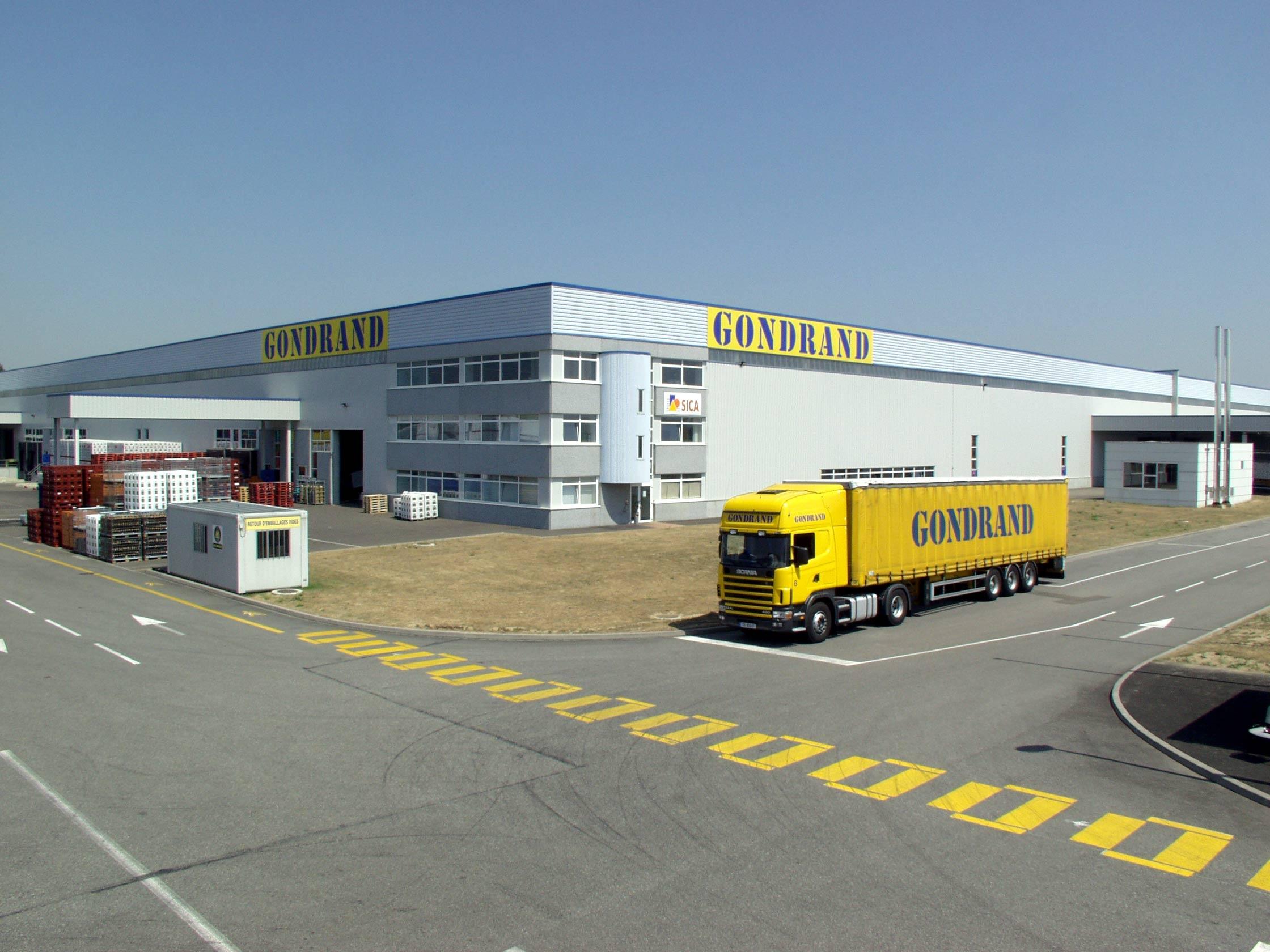}\\
\includegraphics[height=.5\textwidth]{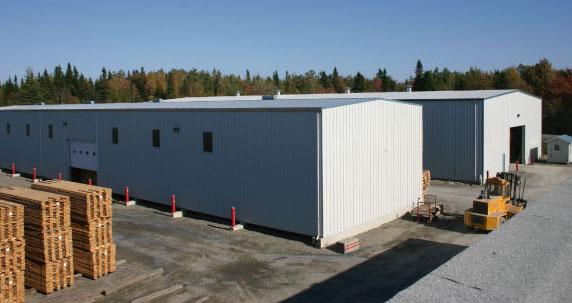}\\
\includegraphics[height=.5\textwidth]{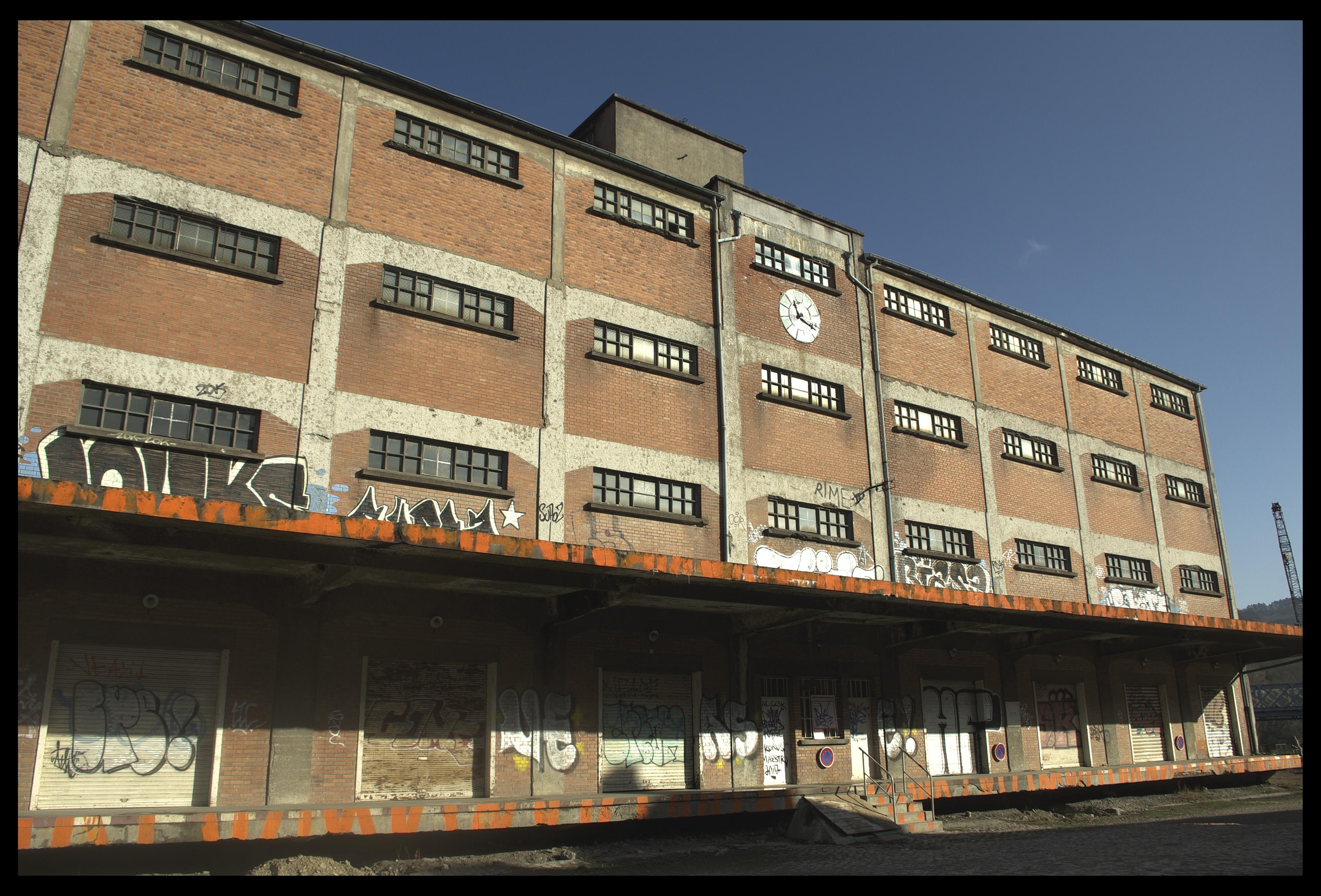}
\end{center}
\end{minipage}\hfill
\begin{minipage}{.45\textwidth}
\begin{center}
\includegraphics[height=.5\textwidth]{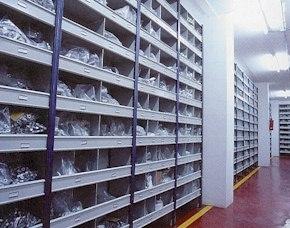}\\
\includegraphics[height=.5\textwidth]{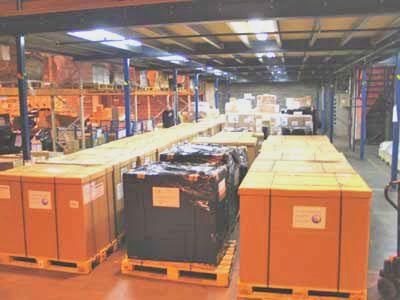}\\
\includegraphics[height=.5\textwidth]{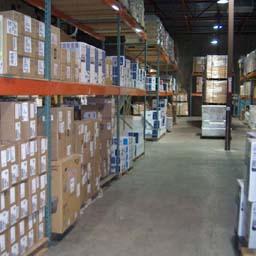}
\end{center}
\end{minipage}
\end{center}
\caption{Test images. Positive examples $p_1$, $p_2$, $p_3$ on the left (from top), negative examples $n_1$, $n_2$, $n_3$ on the right (from top).}\label{fig:warehouse-imgs}
\end{figure}

We utilize the ADE20K dataset \cite{zhou2016semantic,zhou2017scene}. It contains 20,000 images of scenes which have been pre-classified regarding scenes depicted, i.e., we assume that the classification is done by a trained neural network.\footnote{Strictly speaking, this is not true for the training subset of the ADE20K dataset, but that doesn't really matter for our demonstration.}
For our initial test, we used six images, three of which have been classified as ``outdoor warehouse'' scenes (our positive examples), and three of which have not been classified as such (our negative examples). In fact, for simplicity, we took the negative examples from among the images which had been classified as ``indoor warehouse'' scenes. The images are shown in Figure \ref{fig:warehouse-imgs}.

The ADE20K dataset furthermore provides annotations for each image which identify information about objects which have been identified in the image. The annotations are in fact richer than that and also talk about the number of objects, whether they are occluded, and some more, but for our initial experiment we only used presence or absence of an object. To keep the initial experiment simple, we furthermore only used those detected objects which could easily be mapped to our chosen background knowledge, the Suggested Upper Merged Ontology (SUMO).\footnote{\url{http://www.adampease.org/OP/}} Table \ref{tab:warehouse-objects} shows, for each image, the objects we kept. 
\begin{table}
\begin{tabbing}
image $p_n$: \= \kill
image $p_1$: \> road, window, door, wheel, sidewalk, truck, box, building\\
image $p_2$: \> tree, road, window, timber, building, lumber\\
image $p_3$: \> hand, sidewalk, clock, steps, door, face, building, window, road\\
image $n_1$: \> shelf, ceiling, floor\\
image $n_2$: \> box, floor, wall, ceiling, product\\
image $n_3$: \> ceiling, wall, shelf, floor, product
\end{tabbing}
\caption{Objects recorded for each image.}\label{tab:warehouse-objects}
\end{table}
The Suggested Upper Merged Ontology was chosen because it contains many, namely about 25,000 common terms which cover a wide range of domains. At the same time, the ontology arguably structures the terms in a relatively straightforward manner which seemed to simplify matters for our initial experiment. 

In order to connect the annotations to SUMO, we used a single role called ``contains.'' Each image was made an individual in the knowledge base. Furthermore, for each of the object identifying terms in Table \ref{tab:warehouse-objects}, we either identified a corresponding matching SUMO class, or created one and added it to SUMO by inserting it at an appropriate place within SUMO's class hierarchy. 
We furthermore created individuals for each of the object identifying terms, including duplicates, in Table \ref{tab:warehouse-objects}, and added them to the knowledge base by typing them with the corresponding class. Finally, we related each image individual to each corresponding object individual via the ``contains'' role. 

To exemplify -- for the image $p_1$ we added individuals road1, window1, door1, wheel1, sidewalk1, truck1, box1, building1, declared Road(road1), Window(window1), etc., and finally added the ABox statements $\text{contains}(p_1,\text{road1})$, $\text{contains}(p_1,\text{window1})$, etc., to the knowledge base. For the image $p_2$, we added $\text{contains}(p_2,\text{tree2})$, $\text{contains}(p_2,\text{road2})$, etc. as well as the corresponding type declarations Tree(tree2), Road(road2), etc. 

The mapping of the image annotations to SUMO is of course very simple, and this was done deliberately in order to show that a straightforward approach already yields interesting results. As our work progresses, we do of course anticipate that we will utilize more complex knowledge bases and will need to generate more complex mappings from picture annotations (or features) to the background knowledge. 

Finally, we ran DL-Learner on the knowledge base, with the positive and negative examples as indicated. DL-Learner returns 10 solutions, which are listed in Figure \ref{fig:warehouse-solutions}. Of these, some are straightforward from the image annotations, such as (\ref{sol1}), (\ref{sol5}), (\ref{sol8}, (\ref{sol9}) and (\ref{sol10}). Others, such as (\ref{sol2}), (\ref{sol4}), (\ref{sol6}), (\ref{sol7}) are much more interesting as they provide solutions in terms of the background knowledge without using any of the terms from the original annotation. Solution (\ref{sol3}) looks odd at first sight, but is meaningful in the context of the SUMO ontology: SelfConnectedObject is an abstract class which is a direct child of the class Object in SUMO's class hierarchy. Its natural language definition is given as ``A SelfConnectedObject is any Object that does not consist of two or more disconnected parts.'' As such, the class is a superclass of the class Road, which explains why (\ref{sol3}) is indeed a solution in terms of the SUMO ontology.

\begin{figure}[t]
\begin{minipage}{.45\textwidth}
\begin{gather}
\exists\text{contains}.\text{Window}\label{sol1}\\
\exists\text{contains}.\text{Transitway}\label{sol2}\\
\exists\text{contains}.\text{SelfConnectedObject}\label{sol3}\\
\exists\text{contains}.\text{Roadway}\label{sol4}\\
\exists\text{contains}.\text{Road}\label{sol5}
\end{gather}
\end{minipage}
\begin{minipage}{.45\textwidth}
\begin{gather}
\exists\text{contains}.\text{LandTransitway}\label{sol6}\\
\exists\text{contains}.\text{LandArea}\label{sol7}\\
\exists\text{contains}.\text{Building}\label{sol8}\\
\forall\text{contains}.\neg\text{Floor}\label{sol9}\\
\forall\text{contains}.\neg\text{Ceiling}\label{sol10}
\end{gather}
\end{minipage}
\caption{Solutions produced by DL-Learner for the warehouse test.}\label{fig:warehouse-solutions}
\end{figure}

We have conducted four additional experiments along the same lines as described above. We briefly describe them below -- the full raw data and results are available from {\small \url{http://daselab.org/projects/human-centered-big-data}}.

In the second experiment, we chose four workroom pictures as positive examples, and eight warehouse pictures (indoors and outdoors) as negative examples. An example explanation DL-Learner came up with is $$\exists\text{contains}.(\text{DurableGood} \sqcap \neg\text{ForestProduct}).$$ On of the outdoor warehouse pictures indeed shows timber. DurableGoods in SUMO include furniture, machinery, and appliances.

In the third experiment, we chose the same four workroom pictures as negative examples, and the same eight warehouse pictures (indoors and outdoors) as positve examples. An example explanation DL-Learner came up with is 
$$\forall\text{contains}.(\neg \text{Furniture} \sqcap \neg\text{IndustrialSupply}),$$
i.e., ``contains neither furniture nor industrial supply''. IndustrialSupply in SUMO includes machinery. Indeed it turns out that furniture alone is insufficient for distingushing between the positive and negative exaples, because ``shelf'' is not classified as funiture in SUMO. This shows the dependency of the explanations on the conceptualizations encoded in the background knowledge. 

In the fourth experiment, we chose eight market pictures (indoors and outdoors) as positive examples, and eight warehouse pictures (indoors and outdoors) as well as four workroom pictures as negative examples. An example explanation DL-Learner came up with is 
$$\exists\text{contains}.\text{SentientAgent},$$
And indeed it turns out that people are shown on all the market pictures. There is actually also a man shown on one of the warehouse pictures, driving a forklift, however ``man'' or ``person'' was not among the annotations used for the picture. This example indicates how our approach could be utilized: A human monitor inquiring with an interactive system about the reasons for a certain classification may notice that the man was missed by the software on that particular picture, and can opt to interfere with the decision and attempt to correct it. 

In the fifth experiment, we chose four mountain pictures as positive examples, and eight warehouse pictures (indoors and outdoors) as well as four workroom pictures as negative examples. An example explanation DL-Learner came up with is 
$$\exists\text{contains}.\text{BodyOfWater}.$$ Indeed, it turns out that all mountain pictures in the example set show either a river or a lake. Similar to the previous example, a human monitor may be able to catch that some misclassifications may occur because presence of a body of water is not always indicative of presence of a mountain.

%% file: conclusion.tex
\section{Conclusions and Further Work}\label{sec:conclusion}


We have laid out a conceptual sketch how to approach the issue of explaining artificial neural networks' classification behaviour using Semantic Web background knowledge and technologies, in a non-propositional setting. We have also reported on some very preliminary experiments to support our concepts. 

The sketch already indicates where to go from here: We will need to incorporate more complex and more comprehensive background knowledge, and if readily available structured knowledge turns out to be insufficient, then we foresee using state of the art knowledge graph generation and ontology learning methods \cite{JiG11,LehmannV14} to obtain suitable background knowledge. We will need to use automatic methods for mapping network input features to the background knowledge \cite{euzenat-om,UrenCIHVMC06}, while the features to be mapped may have to be generated from the input in the first place, e.g. using object recognition software in the case of images. And finally, we also intend to apply the approach to sets of hidden neurons in order to understand what their activations indicate.

%% file: main.bbl
\begin{thebibliography}{10}
\providecommand{\url}[1]{\texttt{#1}}
\providecommand{\urlprefix}{URL }

\bibitem{BCMNP07}
Baader, F., Calvanese, D., McGuinness, D., Nardi, D., Patel-Schneider, P.F.
  (eds.): The Description Logic Handbook: Theory, Implementation, and
  Applications. Cambridge University Press, 2nd edn. (2010)

\bibitem{BaderH05}
Bader, S., Hitzler, P.: Dimensions of neural-symbolic integration -- {A}
  structured survey. In: Art{\"{e}}mov, S.N., Barringer, H., d'Avila Garcez,
  A.S., Lamb, L.C., Woods, J. (eds.) We Will Show Them! Essays in Honour of Dov
  Gabbay, Volume One. pp. 167--194. College Publications (2005)

\bibitem{turtle}
Beckett, D., Berners-Lee, T., Prud'hommeaux, E., Carothers, G.: {RDF 1.1.
  Turtle -- Terse RDF Triple Language}. {W3C} Recommendation (25 February
  2014), available at http://www.w3.org/TR/turtle/

\bibitem{tbl-semantic-web}
Berners-Lee, T., Hendler, J., Lassila, O.: The {S}emantic {W}eb. Scientific
  American  284(5),  34--43 (May 2001)

\bibitem{lod2009}
Bizer, C., Heath, T., Berners-Lee, T.: {Linked Data -- The Story So Far}.
  International Journal on Semantic Web and Information Systems  5(3),  1--22
  (2009)

\bibitem{BuhmannLW16}
B{\"{u}}hmann, L., Lehmann, J., Westphal, P.: {DL-Learner} -- {A} framework for
  inductive learning on the semantic web. Journal of Web Semantics  39,  15--24
  (2016)

\bibitem{euzenat-om}
Euzenat, J., Shvaiko, P.: Ontology Matching, Second Edition. Springer (2013)

\bibitem{AAAISS15}
Garcez, A., Besold, T., de~Raedt, L., F\"oldiak, P., Hitzler, P., Icard, T.,
  K\"uhnberger, K.U., Lamb, L., Miikkulainen, R., Silver, D.: Neural-symbolic
  learning and reasoning: Contributions and challenges. In: Gabrilovich, E.,
  Guha, R., McCallum, A., Murphy, K. (eds.) Proceedings of the AAAI 2015 Spring
  Symposium on Knowledge Representation and Reasoning: Integrating Symbolic and
  Neural Approaches. Technical Report, vol. SS-15-03. AAAI Press, Palo Alto, CA
  (2015)

\bibitem{AZ99}
d'Avila Garcez, A.S., Zaverucha, G.: The connectionist inductive lerarning and
  logic programming system. Applied Intelligence  11(1),  59--77 (1999)

\bibitem{GuhaBM16}
Guha, R.V., Brickley, D., Macbeth, S.: Schema.org: evolution of structured data
  on the web. Commun. {ACM}  59(2),  44--51 (2016)

\bibitem{owl2-primer}
Hitzler, P., Kr\"o{}tzsch, M., Parsia, B., Patel-Schneider, P.F., Rudolph, S.
  (eds.): {OWL}~2 Web Ontology Language Primer (Second Edition). {W3C}
  Recommendation (11 December 2012), \url{http://www.w3.org/TR/owl2-primer/}

\bibitem{FOST}
Hitzler, P., Kr\"otzsch, M., Rudolph, S.: Foundations of Semantic Web
  Technologies. CRC Press/Chapman \&{} Hall (2010)

\bibitem{JiG11}
Ji, H., Grishman, R.: Knowledge base population: Successful approaches and
  challenges. In: Lin, D., Matsumoto, Y., Mihalcea, R. (eds.) The 49th Annual
  Meeting of the Association for Computational Linguistics: Human Language
  Technologies, Proceedings of the Conference, 19-24 June, 2011, Portland,
  Oregon, {USA}. pp. 1148--1158. The Association for Computer Linguistics
  (2011)

\bibitem{nesy17-prop}
Labaf, M., Hitzler, P., Evans, A.B.: Propositional rule extraction from neural
  networks under background knowledge. In: Proceedings of the Twelfth
  International Workshop on Neural-Symbolic Learning and Reasoning, NeSy'17,
  London, UK, July 2017 (2017), to appear

\bibitem{LarsonM77}
Larson, J., Michalski, R.S.: Inductive inference of {VL} decision rules.
  {SIGART} Newsletter  63,  38--44 (1977)

\bibitem{LehmannBH10}
Lehmann, J., Bader, S., Hitzler, P.: Extracting reduced logic programs from
  artificial neural networks. Applied Intelligence  32(3),  249--266 (2010)

\bibitem{LehmannH10}
Lehmann, J., Hitzler, P.: Concept learning in description logics using
  refinement operators. Machine Learning  78(1-2),  203--250 (2010)

\bibitem{LehmannIJJKMHMK15}
Lehmann, J., Isele, R., Jakob, M., Jentzsch, A., Kontokostas, D., Mendes, P.N.,
  Hellmann, S., Morsey, M., van Kleef, P., Auer, S., Bizer, C.: {DB}pedia --
  {A} large-scale, multilingual knowledge base extracted from {W}ikipedia.
  Semantic Web  6(2),  167--195 (2015)

\bibitem{LehmannV14}
Lehmann, J., V{\"{o}}lker, J.: Perspectives on Ontology Learning, Studies on
  the Semantic Web, vol.~18. {IOS} Press (2014)

\bibitem{MuggletonR94}
Muggleton, S., Raedt, L.D.: Inductive logic programming: Theory and methods.
  Journal of Logic Programming  19/20,  629--679 (1994)

\bibitem{UrenCIHVMC06}
Uren, V.S., Cimiano, P., Iria, J., Handschuh, S., Vargas{-}Vera, M., Motta, E.,
  Ciravegna, F.: Semantic annotation for knowledge management: Requirements and
  a survey of the state of the art. J. Web Sem.  4(1),  14--28 (2006)

\bibitem{VrandecicK14}
Vrandecic, D., Kr{\"{o}}tzsch, M.: Wikidata: a free collaborative
  knowledgebase. Commun. {ACM}  57(10),  78--85 (2014)

\bibitem{zhou2016semantic}
Zhou, B., Zhao, H., Puig, X., Fidler, S., Barriuso, A., Torralba, A.: Semantic
  understanding of scenes through the {ADE20K} dataset. arXiv preprint
  arXiv:1608.05442  (2016)

\bibitem{zhou2017scene}
Zhou, B., Zhao, H., Puig, X., Fidler, S., Barriuso, A., Torralba, A.: Scene
  parsing through {ADE20K} dataset. In: Proceedings of the IEEE Conference on
  Computer Vision and Pattern Recognition (2017)

\end{thebibliography}
